\pdfoutput=1
\documentclass{article}
\usepackage{spconf,amsmath,graphicx}
\usepackage{stfloats}
\usepackage{balance}
\usepackage[hidelinks,pdfusetitle]{hyperref}
\hypersetup{pdfauthor={Andreas Schwarz, Christian Huemmer, Roland Maas, Walter Kellermann},pdftitle={Spatial Diffuseness Features for DNN-Based Speech Recognition in Noisy and Reverberant Environments}}

\usepackage{url}
\usepackage[utf8]{inputenc}

\usepackage{tikz}

\usepackage{pgfplots}
\usepgfplotslibrary{groupplots}
\pgfplotsset{compat=newest}
\pgfplotsset{plot coordinates/math parser=false}
\newlength\figureheight
\newlength\figurewidth

\renewcommand{\normalsize}{\fontsize{9.38pt}{11.38pt}\selectfont}

\newcounter{MYtempeqncnt}

\renewcommand\Re{\operatorname{Re}}

\newcommand{\frameix}{\ensuremath{k}}

\newcommand{\CDR}{\ensuremath{\mathit{CDR}}}

\newcommand{\SNR}{\ensuremath{\mathit{SNR}}}

\newcommand{\sinc}{\ensuremath{\operatorname{sinc}}}

\title{Spatial Diffuseness Features for DNN-Based Speech Recognition\\in Noisy and Reverberant Environments}
\name{Andreas Schwarz, Christian Huemmer, Roland Maas, Walter Kellermann\thanks{The authors would like to thank the Deutsche Forschungsgemeinschaft (DFG) for supporting this work (contract number KE 890/4-2).}}
\address{Multimedia Communications and Signal Processing\\Friedrich-Alexander-Universität Erlangen-Nürnberg (FAU)\\Cauerstr. 7, 91058 Erlangen, Germany\\ {\tt \small \{schwarz, huemmer, maas, wk\}@lnt.de}}

\begin{document}
\maketitle
\begin{abstract}
We propose a spatial diffuseness feature for deep neural network (DNN)-based automatic speech recognition to improve recognition accuracy in reverberant and noisy environments. The feature is computed in real-time from multiple microphone signals without requiring knowledge or estimation of the direction of arrival, and represents the relative amount of diffuse noise in each time and frequency bin. It is shown that using the diffuseness feature as an additional input to a DNN-based acoustic model leads to a reduced word error rate for the REVERB challenge corpus, both compared to logmelspec features extracted from noisy signals, and features enhanced by spectral subtraction.
\end{abstract}
\begin{keywords}
Speech Recognition, Reverberation, Diffuse Noise, Deep Neural Networks
\end{keywords}
\section{Introduction}
\label{sec:intro}

\begin{figure*}[b!]
\hrulefill
\vspace{-1mm}
\normalsize
\setcounter{MYtempeqncnt}{\value{equation}}
\setcounter{equation}{7}
\begin{equation}
\label{eq:cdr-estimator}
\widehat{\CDR}(k,f) = \frac{\Gamma_n\, \Re\{\hat\Gamma_x\} -{|\hat\Gamma_x|}^2 - \sqrt{\Gamma_n^2\, {\Re\{\hat\Gamma_x\}}^2 - \Gamma_n^2\, {|\hat\Gamma_x|}^2 + \Gamma_n^2 - 2\, \Gamma_n\, \Re\{\hat\Gamma_x\} + {|\hat\Gamma_x|}^2}}{{|\hat\Gamma_x|}^2 - 1}
\end{equation}
\setcounter{equation}{\value{MYtempeqncnt}}
\end{figure*}

In automatic speech recognizers (ASR) based on Gaussian Mixture Models and Hidden Markov Models (GMM-HMM), a wide variety of transformations and feature extraction steps is currently being employed with the aim of extracting and normalizing the information contained in the time-domain input signal as efficiently as possible. Recently, with the development of effective training methods for acoustic models based on multiple-layer neural networks, which are often summarized under the term ``deep neural networks'' (DNN) \cite{hinton_deep_2012}, it has become possible for the acoustic model to learn relationships between features and phonemes to a higher degree than it is possible with manually implemented feature transformation steps. For example, it has been found that simple filterbank features outperform mel-frequency cepstral coefficients (MFCCs) \cite{deng_recent_2013,seltzer_investigation_2013}, and it is conceivable that, given large amounts of training data and sufficiently complex network structures, time-domain signals may at some point even be directly used as inputs to a neural network.

Although the trend in ASR goes towards replacing explicit processing stages by implicit learning, for noise- and reverberation-robust ASR using microphone arrays, spatial information is still predominantly being exploited in a separate speech enhancement preprocessor, e.g., in the form of beamforming \cite{weninger_merl/melco/tum_2014}, multichannel linear prediction \cite{delcroix_linear_2014}, blocking matrix-based postfilters \cite{maas_two-channel_2011} or coherence-based postfilters \cite{schwarz_coherent-to-diffuse_2015}. The single-channel output of the preprocessor is then used to compute features for ASR. In some GMM-HMM-based systems, spatial information is exploited indirectly in uncertainty decoding-based approaches, e.g., in \cite{astudillo_integration_2013}, where the feature uncertainty is derived from a noise estimate obtained in a multichannel signal enhancement stage. For DNN-based acoustic models, ``noise-aware training'' has been proposed \cite{seltzer_investigation_2013}, where a noise estimate is appended to the noisy feature vector. This has been evaluated for stationary noise estimates \cite{seltzer_investigation_2013} and noise-estimates derived from time-frequency masking \cite{narayanan_joint_2014}, but may in principle also be used for noise estimates obtained from spatial processing. In \cite{swietojanski_hybrid_2013} and \cite{liu_using_2014}, feature vectors from multiple microphones are concatenated to form the input of a DNN-based acoustic model, however, no spatial phase information is exploited.

Inspired by the trend towards moving more explicit feature processing steps into the DNN, we propose to exploit spatial information about the diffuseness of the sound field directly by incorporating it into the acoustic model of a DNN-based speech recognizer. The diffuseness estimate is derived from the complex coherence between two omnidirectional microphones and has been used for signal enhancement based on the assumption that late reverberation and noise components can be modeled as diffuse noise \cite{schwarz_coherent-to-diffuse_2015}. Using the diffuseness as a feature is motivated by the fact that humans exploit similar spatial information for speech recognition in reverberant and noisy environments \cite{danilenko_binaurales_1969,culling_speech_2006}, as it was found that the human auditory system treats spectro-temporal variations in the interaural coherence as ``a perceptual surrogate for spectro-temporal variations in the energy of speech signals'' \cite{culling_speech_2006}. The aim is to learn similar behavior in a DNN-based acoustic model.

We first describe the signal model for the estimation of the diffuseness from the instantaneous spatial coherence of a reverberated and noisy speech signal. Then, we show how this estimate is integrated into a feature extraction scheme for ASR, and describe the structure of the DNN-based speech recognizer. Finally, we evaluate the proposed feature on the two-channel task of the REVERB challenge \cite{kinoshita_reverb_2013}, showing that the proposed approach outperforms both noisy multi-condition training and multichannel spectral subtraction-based signal enhancement.

\begin{figure*}[t!]
\centering
\includegraphics[width=14cm]{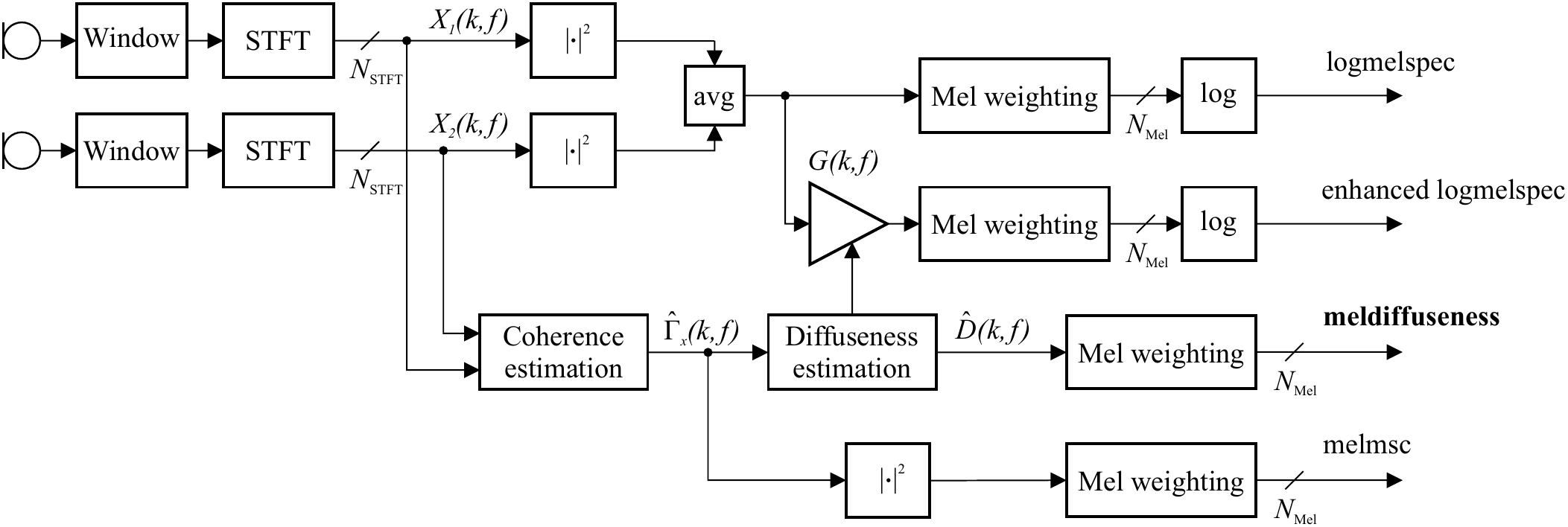}
\caption{Feature extraction of logmelspec, enhanced logmelspec, meldiffuseness and melmsc features from 2-channel signals.}
\label{fig:feature-extraction}
\end{figure*}

\section{Blind Diffuseness Estimation}
\label{sec:format}

We consider a reverberated and noisy speech signal recorded by two omnidirectional microphones. The signal $x_i(t)$ recorded at the $i$-th microphone is composed of the desired signal component $s_i(t)$ and the undesired noise component $n_i(t)$ comprising additive noise and late reverberation, i.e.,
$x_i(t) = s_i(t) + n_i(t),~i=1,2.$
The microphone, desired, and noise signals are represented in the short-time Fourier transform (STFT) domain by the corresponding uppercase letters, i.e., $X_i(\frameix,f)$, $S_i(\frameix,f)$ and $N_i(\frameix,f)$, respectively, with the discrete frame index $\frameix$ and continuous frequency $f$, and the auto- and cross-power spectra $\Phi_{x_i x_j}(\frameix,f)$, $\Phi_{s_i s_j}(\frameix,f)$, $\Phi_{n_i n_j}(\frameix,f)$. Note that the continuous frequency $f$ is used here for generality; in practice, $f$ denotes discrete values along the frequency axis.
It is assumed that the auto-power spectra of all signal components are identical at both microphones, i.e.,
$\Phi_{s_i s_i}(\frameix,f)=\Phi_{s}(\frameix,f), \Phi_{n_i n_i}(\frameix,f) = \Phi_{n}(\frameix,f)$.
The time- and frequency-dependent signal-to-noise ratio (SNR) of the microphone signals can then be defined as
\begin{flalign}
\SNR(\frameix,f)=\frac{\Phi_{s}(\frameix,f)}{\Phi_{n}(\frameix,f)}.
\end{flalign}
The complex spatial coherence functions of the desired signal and noise components are given by
\begin{flalign}
\Gamma_s(f)=\frac{\Phi_{s_1 s_2}(\frameix,f)}{\Phi_{s}(\frameix,f)},
\Gamma_n(f)=\frac{\Phi_{n_1 n_2}(\frameix,f)}{\Phi_{n}(\frameix,f)},
\end{flalign}
and are assumed to be time-invariant, i.e., dependent only on the spatial characteristics of the signal components. It is furthermore assumed that signal and noise components are orthogonal, such that
$\Phi_x(\frameix,f) = \Phi_s(\frameix,f) + \Phi_n(\frameix,f).$
The complex spatial coherence of the mixed sound field can then be written as a function of the SNR and the signal and noise coherence functions:
\begin{flalign}
\Gamma_{x}(\frameix,f) = \frac{ \SNR(\frameix,f) \Gamma_{s}(f) + \Gamma_{n}(f) }{ \SNR(\frameix,f) + 1 }.
\label{eq:Gamma_x}
\end{flalign}
The direct sound is now modeled as a plane wave with an unknown direction of arrival (DOA) and therefore unknown time difference of arrival $\Delta t$, while the undesired noise and late reverberation component is modeled as a diffuse (spherically isotropic) sound field. The corresponding spatial coherence functions for the direct and diffuse sound components are then given by
\begin{flalign}
\label{eq:coherencemodel}
\Gamma_s(f) &= e^{j 2 \pi f \Delta t}, \\
\Gamma_n(f) &= \Gamma_\text{diffuse}(f) = \sinc(2 \pi f \frac{d}{c}),
\end{flalign}
respectively. The direct signal coherence has a magnitude of one with an unknown phase determined by the DOA, while the diffuse noise coherence only depends on the known microphone spacing $d$.

The aim in the following is to estimate the SNR from the coherence of the mixed sound field $\Gamma_x(\frameix,f)$. This coherence is first estimated as
\begin{equation}
\label{eq:coherence}
\hat\Gamma_x(\frameix,f)=\frac{\hat\Phi_{x_1 x_2}(\frameix,f)}{\sqrt{\hat\Phi_{x_1 x_1}(\frameix,f) \hat\Phi_{x_2 x_2}(\frameix,f)}},
\end{equation}
where the spectral estimates $\hat\Phi_{x_i x_j}(\frameix,f)$ are obtained by recursive averaging:
\begin{equation}
\label{eq:recursive}
\hat\Phi_{x_i x_j}(\frameix,f)=\lambda \hat\Phi_{x_i x_j}(\frameix\!-\!1,f) + (1\!-\!\lambda) X_i(\frameix,f) X_j^*(\frameix,f),
\end{equation}
with a constant forgetting factor $\lambda$ between 0 and 1.
In \cite{schwarz_unbiased_2014,schwarz_coherent-to-diffuse_2015}, it was shown that (\ref{eq:Gamma_x}) can be solved for the SNR without requiring knowledge of $\Gamma_s$, using only the assumption that the desired signal is fully coherent, i.e., $|\Gamma_s|=1$. This yields a ``blind'' estimator for the SNR (or coherent-to-diffuse ratio, CDR) from the mixture coherence $\hat\Gamma_x(\frameix,f)$ which does not require knowledge or estimation of the signal DOA. The estimator is given in (\ref{eq:cdr-estimator}) at the bottom of this page (the indices $k$ and $f$ are omitted for brevity).%
\addtocounter{equation}{1} 
The CDR can be transformed into the diffuseness \cite{del_galdo_diffuse_2012}
\begin{flalign}
\label{eq:diffuseness}
\hat D(\frameix,f) = [\widehat\CDR(\frameix,f)+1]^{-1},
\end{flalign}
which can be thought of as the relative amount of diffuse signal power in the respective time and frequency bin. Since the diffuseness is bounded between 0 and 1, it is more convenient to use as basis for feature computation than the CDR itself.

\section{Feature Extraction for ASR}
\label{sec:feature-extraction}

Fig.~\ref{fig:feature-extraction} shows the block diagram of the proposed feature extraction scheme. The microphone signals are first windowed and transformed into the STFT domain. The upper path then corresponds to a classical feature extraction of $N_\mathrm{Mel}$-dimensional logmelspec (often termed ``log-filterbank'' or ``Log FBank'') features, where the two microphone signals are combined by averaging the spectral powers computed from each microphone, and $N_\mathrm{Mel}$ triangular Mel-scaled weighting filters are applied. The second path shows the extraction of enhanced logmelspec features, where signal enhancement based on the diffuseness estimate is performed by multiplication in the STFT domain with a gain factor $G(\frameix,f)$, which is computed as described in \cite{schwarz_coherent-to-diffuse_2015} according to the spectral magnitude subtraction rule. The third path illustrates the computation of the proposed ``meldiffuseness'' features: the diffuseness $\hat D(\frameix,f)$ is estimated as described in the previous section, and the same $N_\mathrm{Mel}$ triangular Mel weighting filters that are used in the logmelspec feature extraction are applied to create an output vector of the dimensionality $N_\mathrm{Mel}$. Finally, for comparison, the Mel-weighted magnitude-squared coherence (``melmsc'') is computed as a feature. While the magnitude-squared coherence of a mixed sound field is also related to the amount of diffuse noise, this relationship is strongly dependent on the signal DOA and the microphone spacing, therefore the melmsc feature is expected to perform worse than the proposed diffuseness estimate.

The interesting question is now how using concatenated logmelspec and meldiffuseness features as input to the neural network compares to using logmelspec features which have been enhanced in the STFT domain.

Since the trend in DNN-based acoustic modeling goes towards replacing explicit feature preprocessing and normalization steps by implicit learning, one might consider using the complex spatial coherence directly as feature. Note, however, that the proposed diffuseness feature has two significant advantages over the complex coherence. The complex coherence depends on two additional variables, namely the DOA and the microphone spacing, both of which would need to be sufficiently represented in the training data. Moreover, the diffuseness is a characteristic of the sound field which is independent of the microphone array geometry, and may therefore also be estimated from microphone arrays with other geometries, e.g., spherical arrays \cite{jarrett_coherence-based_2012} or arrays consisting of directional microphones \cite{thiergart_power-based_2014}, without requiring adaptation of the acoustic model.

It is interesting to note that the additional temporal smoothing which is required for the estimation of the coherence (and therefore the diffuseness) has parallels in the human auditory system, where reaction to changes in interaural coherence was found to be more sluggish than reaction to changes in energy \cite{culling_binaural_2000}.

For the results presented in this paper, the time-domain signals (sampled at 16 kHz) are windowed using a 25\,ms Hann window with a frame shift of 10\,ms and transformed using a 512-point DFT, resulting in $N_\mathrm{STFT}=257$ subbands in the STFT domain. The spatial coherence is estimated using the forgetting factor $\lambda=0.68$. $N_\mathrm{Mel}=24$ triangular Mel-scale weighting filters are used, covering a frequency range from 64 to 8000\,Hz. MATLAB code for the feature computation is provided online\footnote{\url{http://www.lms.lnt.de/files/publications/icassp2015-diffuseness.zip}}.

\begin{figure}[tb]
\centering
	\setlength\figureheight{1.4cm}
	\setlength\figurewidth{5.6cm}
	\pgfplotsset{
	title style={font=\footnotesize},
	tick label style={font=\scriptsize},
	label style={font=\scriptsize},
	legend style={font=\scriptsize},
	}
	\input{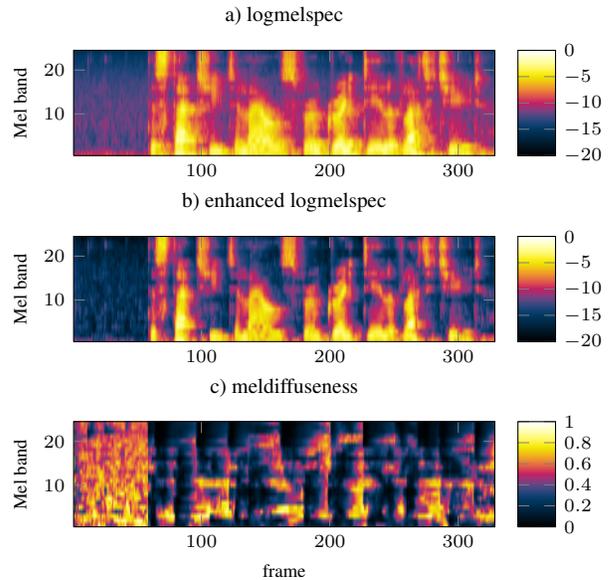}
    \caption{Features for the reverberated utterance ``The statute allows for a great deal of latitude''.}
    \label{fig:features}
\end{figure}

Fig.~\ref{fig:features} illustrates the features computed from a noisy and reverberated speech signal taken from the multi-condition training set of the REVERB challenge corpus (LargeRoom2). The coherence-based spectral enhancement visibly reduces the noise floor and the smearing of the speech features over time. The meldiffuseness clearly highlights portions of the signal where noise or reverberation components are dominant.

\begin{table*}[t]
\centering
\caption{ASR Word Error Rate for the REVERB challenge evaluation and development test sets.}\vspace{2mm}
\includegraphics[width=\textwidth]{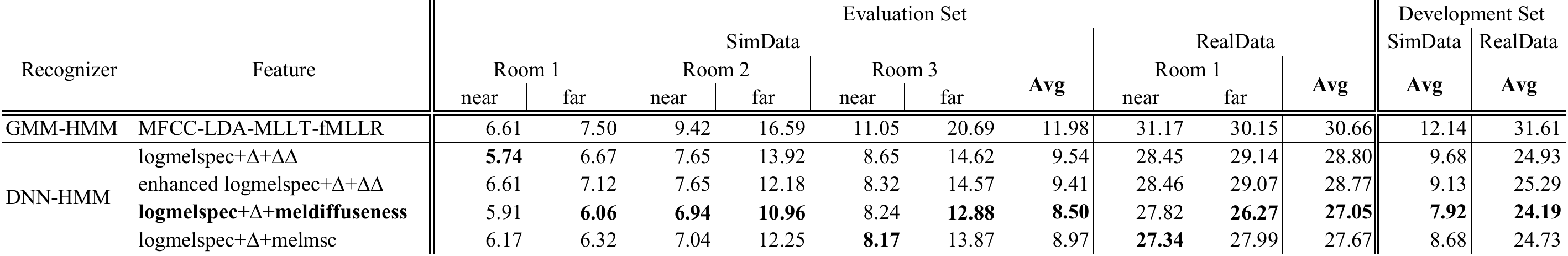}
\vspace{-3mm}
\label{table:results}
\end{table*}

\section{DNN-based Speech Recognition}
\label{sec:dnn}

We employ the Kaldi toolkit \cite{povey_kaldi_2011} as ASR back-end system using the WSJ0 trigram 5k language model of the REVERB challenge and 3551 context-dependent triphone-states in the acoustic model.

In a first step, we set up a GMM-HMM baseline system based on Weninger et al. \cite{weninger_merl/melco/tum_2014}
by extracting 13 mean and variance normalized MFCCs (including the zeroth cepstral coefficient), followed by $\pm 4$ frame splicing, linear discriminant analysis (LDA), maximum likelihood linear transform (MLLT),
and feature-space maximum likelihood linear regression (fMLLR) (see \cite{weninger_merl/melco/tum_2014,rath_2013} for a detailed description).
After conventional maximum likelihood training, discriminative training is performed with the boosted maximum mutual information (bMMI) criterion \cite{weninger_merl/melco/tum_2014}.
The GMM-HMM system is trained on the clean WSJCAM0 Cambridge Read News REVERB corpus \cite{robinson_wsjcam0_2013}.
The alignment of the training data to the HMM states is then extracted from the clean training data and used for the later multi-condition training of the DNN-HMM system.
This technique is known to yield better results than a multi-condition state-frame alignment \cite{narayanan_joint_2014, delcroix_is_2013}.

The hybrid DNN-HMM Kaldi system is based on ``Dan's implementation'' \cite{povey_kaldi_2011}
using a maxout network with 2-norm nonlinearities/activation functions and 4 hidden layers, each one with an input dimension of 2000 and an output dimension of 400.

In accordance with \cite{deng_recent_2013, seltzer_investigation_2013}, and as described in the previous section, we extract $N_\mathrm{Mel}=24$ static logmelspec coefficients, with or without applying coherence-based spectral subtraction enhancement in the STFT domain.
Depending on the particular setup in Table~\ref{table:results}, also Delta ($\Delta$), acceleration ($\Delta\Delta$), melmsc, and/or the proposed meldiffuseness features are derived. Mean and variance normalization and $\pm 5$ frame splicing is applied to the entire resulting feature vector.
The training is performed on the REVERB multi-condition training set \cite{kinoshita_reverb_2013}, consisting of 7861 noisy and reverberated utterances from the WSJCAM0 corpus, using greedy layer-wise supervised training, preconditioned stochastic gradient descent, ``mixing up'' \cite{zhang_improving_2014} as well as final model combination \cite{zhang_improving_2014}.

\section{Evaluation Results}
\label{sec:evaluation}
We evaluate the proposed system using the two-channel task of the REVERB challenge \cite{kinoshita_reverb_2013}.
The REVERB evaluation test set consists of $\sim$5000 reverberated and noisy utterances, partially created by convolution of clean WSJCAM0 utterances with impulse responses and mixing with recorded noise sequences (``SimData''), and partially consisting of multichannel recordings of speakers in a reverberant and noisy room from the MC-WSJ-AV corpus (``RealData''). For SimData, the reverberation times of the three rooms are approx. 0.25\,s, 0.5\,s and 0.7\,s and the source-microphone spacing is 0.5\,m (near) or 2\,m (far). For RealData, the reverberation time is approx 0.7\,s and the source-microphone distance is 1\,m (near) or 2.5\,m (far). In both cases, an 8-channel circular microphone array with a diameter of 20\,cm was used, of which two microphones with a spacing of $d=8\,\mathrm{cm}$ are selected for the two-channel recognition task which is evaluated here.

First, we evaluate the word error rate (WER) obtained from the GMM-based recognizer with MFCC features, which is used to obtain the alignment. For the DNN-based recognizer, we compare logmelspec features extracted from the noisy signals, and enhanced logmelspec features. In both cases, the feature vector is extended by first- ($\Delta$) and second-order ($\Delta\Delta$) derivatives. Then, we evaluate the combination of noisy logmelspec features with spatial meldiffuseness or melmsc features; in this case, only first-order derivatives ($\Delta$) are computed for the logmelspec features, in order to keep the overall dimension of the feature vectors the same ($3 N_\mathrm{Mel}$).

Table~\ref{table:results} shows the WER results for the REVERB challenge evaluation test set, and the average WER for the development test set. As expected, the DNN-based acoustic model achieves a lower WER than the GMM-based model. The diffuseness-based signal enhancement has a negligible effect on WER. This seems to contradict \cite{schwarz_unbiased_2014}, where the same signal enhancement method led to a significantly lower WER, however, there, acoustic models were trained on clean speech. Apparently the effect of the multichannel spectral subtraction for signal enhancement is compensated by noisy multi-condition training. Using the combined noisy logmelspec and diffuseness features as input to the neural network however yields a significantly reduced WER. This confirms that the spatial information extracted from the coherence can be exploited more successfully by the DNN than by speech enhancement using spectral subtraction, even though, in this case, the frequency resolution of the meldiffuseness features is reduced compared to the diffuseness estimate used for spectral subtraction. The melmsc feature also leads to a reduced WER compared to noisy logmelspec features, although the improvement is smaller than with meldiffuseness features.

\section{Conclusion}
\label{sec:conclusion}

It has been shown that spatial information extracted from multiple microphones does not necessarily have to be exploited in a signal enhancement front-end, but may be used more effectively as an additional feature input for a DNN-based speech recognizer. The proposed approach has a number of properties which make it highly suitable for practical applications like cloud-based speech recognition for smartphones. First, the diffuseness feature is normalized with respect to the microphone array geometry, and can therefore be used for speech recognition with features extracted from a variety of multichannel recording devices without requiring adaptation of the acoustic model. Second, the feature can be computed in real-time (as opposed to batch processing) and ``blindly'' in the sense that knowledge or estimation of the direction of arrival is not required. Finally, the evaluation shows that consistent improvements in recognition accuracy can be achieved.

\pagebreak
\balance
\bibliographystyle{IEEEbib}
\bibliography{literature}

\end{document}